\documentclass[conference]{IEEEtran}
\IEEEoverridecommandlockouts
\usepackage{cite}
\usepackage{amsmath,amssymb,amsfonts}
\usepackage{algorithmic}
\usepackage{graphicx}
\usepackage{textcomp}
\usepackage{xcolor}
\usepackage{float}
\usepackage[normalem]{ulem}
\useunder{\uline}{\ul}{}
\def\BibTeX{{\rm B\kern-.05em{\sc i\kern-.025em b}\kern-.08em
    T\kern-.1667em\lower.7ex\hbox{E}\kern-.125emX}}
\begin{document}

\title{Mixer is more than just a model\\
%
}



\author{
	\IEEEauthorblockN{
		1\textsuperscript{st} Qingfeng Ji\IEEEauthorrefmark{1}, 
		2\textsuperscript{nd} Yuxin Wang\IEEEauthorrefmark{1}, 
		3\textsuperscript{rd} Letong Sun\IEEEauthorrefmark{1}, 
            } 
	\IEEEauthorblockA{\IEEEauthorrefmark{1}School of Computer Science and Technology, Dalian University of Technology, Dalian, PRC}
	\IEEEauthorblockA{1\textsuperscript{st} Email: 15640414255@mail.dlut.edu.cn}
      \IEEEauthorblockA{2\textsuperscript{nd} Email: wyx@dlut.edu.cn, corresponding author}
      \IEEEauthorblockA{3\textsuperscript{rd} Email: 3038847812@mail.dlut.edu.cn}
} 


\maketitle

\begin{abstract}
Recently, MLP structures have regained popularity, with MLP-Mixer standing out as a prominent example. In the field of computer vision, MLP-Mixer is noted for its ability to extract data information from both channel and token perspectives, effectively acting as a fusion of channel and token information. Indeed, Mixer represents a paradigm for information extraction that amalgamates channel and token information. The essence of Mixer lies in its ability to blend information from diverse perspectives, epitomizing the true concept of "mixing" in the realm of neural network architectures. Beyond channel and token considerations, it is possible to create more tailored mixers from various perspectives to better suit specific task requirements. This study focuses on the domain of audio recognition, introducing a novel model named Audio Spectrogram Mixer with Roll-Time and Hermit FFT (ASM-RH) that incorporates insights from both time and frequency domains. Experimental results demonstrate that ASM-RH is particularly well-suited for audio data and yields promising outcomes across multiple classification tasks. The models and optimal weights files will be published.
\end{abstract}

\begin{IEEEkeywords}
Audio Classification, MLP-Mixer, Audio Spectrogram Mixer, FFT, RollBlock
\end{IEEEkeywords}

\section{Introduction}
In the realm of deep learning, Transformers have emerged as dominant players across various domains, exemplified by ViT\cite{2020An} in computer vision and the refined versions like DeiT\cite{2020Training}, Swin Transformer\cite{2021Swin}, and Swin Transformer V2\cite{2021SwinV2}. The rise of ChatGPT further solidifies the Transformer's influence. Yet, the resource-intensive nature of Transformers has prompted a reevaluation of their intricate architectures. Google's MLP-Mixer\cite{2021MLP} presents a compelling case, showcasing that pure MLP structures can rival Transformers in computer vision. Similarly, gMLP\cite{2021Pay} and ResMLP\cite{2021ResMLP} have delivered impressive results, underscoring the efficacy of alternative approaches. Furthermore, the introduction of MTS-Mixer\cite{li2023mts} by the Huawei team for multivariate temporal prediction has demonstrated remarkable capabilities across diverse tasks. Weihao Yu et al. extended this concept by introducing the MetaFormer model\cite{2021MetaFormer}, emphasizing that deep learning models featuring a MetaFormer macro-architecture possess significant potential across a wide spectrum of computer vision tasks. Their proposed PoolFormer outperformed several prominent models, including ResNet\cite{7780459}, ViT\cite{2020An}, DeiT\cite{2020Training}, and Swin Transformer\cite{2021Swin}\cite{2021SwinV2}, in the ImageNet-1K\cite{2015ImageNet} benchmark. Guangting Wang's integration of the Shift operation\cite{wang2022shift}, replacing the attention module with no additional FLOP or parameters, provides a slight edge over the benchmark model Swin Transformer\cite{2021Swin}\cite{2021SwinV2}.

Yuan Gong et al. ventured into the realm of audio classification by introducing the Audio Spectrogram Transformer\cite{gong2021ast}, yielding notable outcomes. They later advanced this work by proposing the SSAST model in an unsupervised fashion\cite{gong2022ssast}. Furthermore, Jiu Feng et al. introduced FlexiAST\cite{Feng2023FlexiASTFI} as an extension of AST, enhancing the model's adaptability by offering flexibility in patch sizes.

Qingfeng Ji et al. introduced the Audio Spectrogram Mixer\cite{Ji2024ASMAS} by merging the AST model with MLP-Mixer and integrating the RGB to grayscale mapping formula. This innovative approach demonstrated outstanding performance across three datasets: SpeechCommands (for audio classification), UrbanSound8K (for environmental classification), and CASIA Chinese Sentiment Corpus (for audio sentiment classification). This introduction represents a significant milestone in bringing the Mixer methodology into the domain of audio classification.

In this study, drawing inspiration from the Shift operation\cite{wang2022shift}\cite{Wu_Wan_Yue_Jin_Zhao_Golmant_Gholaminejad_Gonzalez_Keutzer_2018}\cite{Jeon_Kim_2018}\cite{Yu_Xu_Cai_Sun_Li_2021} in ShiftViT, we introduce the RollBlock module, leading to the development of the Roll-Time-mixing module aimed at enhancing the capture of time-domain information within the speech graph. Furthermore, leveraging the Hermit property for frequency domain information extraction, we introduce the Hermit-Frequency-mixing module. These components are seamlessly integrated with the Audio Spectrogram Mixer, resulting in the creation of the Audio Spectrogram Mixer with Roll-Time and Hermit FFT (ASM-RH). ASM-RH embodies the essence of the Mixer concept by shifting away from the conventional computer vision approach of analyzing spectrograms through Channel and Token perspectives. Instead, ASM-RH adopts a perspective that views spectrogram data from the time and frequency domain angles, aligning more closely with the requirements of audio applications. ASM-RH delivers impressive outcomes across three tasks - SpeechCommand\cite{warden2018speech}, UrbanSound8K\cite{jaiswal2018sound}, and CASIA Chinese Emotion Corpus\cite{ke2018speech}, surpassing ERANNs\cite{verbitskiy2022eranns} to establish a new state-of-the-art (SOTA) performance in the RAVDESS\cite{10.1371/journal.pone.0196391} audio classification task solely based on audio data.

\section{ASM-RH Model}
The structure of Audio Spectrogram Mixer with Roll-Time and Hermit FFT is shown in Fig. 1. The spectrogram data is fed into RH-MixerBlocks after slicing and projection, and then output using an MLP output layer.
\begin{figure}[t]
\centerline{\includegraphics{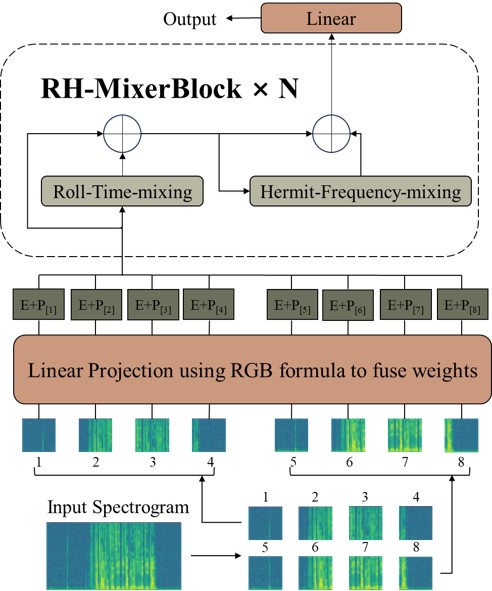}}
\caption{Audio Spectrogram Mixer with Roll-Time and Hermit FFT}
\label{fig1}
\end{figure}
\subsection{Removal of useless structures}

Qingfeng Ji et al., in their proposal of the ASM model\cite{Ji2024ASMAS}, aimed to showcase that Mixer possesses capabilities on par with Transformer in the realm of audio processing. As a result, many of the structures from the AST model\cite{gong2021ast} were retained in the ASM model, although they appeared to be non-essential. Conversely, Ilya Tolstikhin et al., when introducing the MLP-Mixer\cite{2021MLP}, emphasized that the Mixer architecture eliminates the need for positional embedding. Therefore, in ASM-RH, the cls-token and dist-token were omitted accordingly.

\subsection{RollBlock and Roll-Time-mixing}

The ShiftViT\cite{wang2022shift}\cite{Wu_Wan_Yue_Jin_Zhao_Golmant_Gholaminejad_Gonzalez_Keutzer_2018}\cite{Jeon_Kim_2018}\cite{Yu_Xu_Cai_Sun_Li_2021}  proposed by Guangting Wang et al. leverages the Shift operation to efficiently extract information without adding to the parameter complexity or computational load. The Shift operation involves spatially shifting the data matrix by a small increment in four directions while keeping the remaining channels unchanged, and filling the empty spaces with zeros. This operation bears resemblance to convolution. As every segment of time-domain information in the spectrogram is crucial, to maintain the integrity of the time-domain data, the discarded data from the Shift operation is reinstated in the empty positions. This process resembles rolling the data in place, leading to the term RollBlock. As the module's depth increases, we limit the range and distance over which the data scrolls. As shown in Fig. 2.
\begin{figure}[ht]
\centerline{\includegraphics{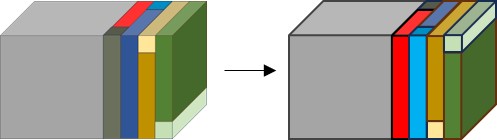}}
\caption{RollBlock}
\label{fig2}
\end{figure}

In detail, the data initially enters the RollBlock in the form of (BatchSize, Height, Width). It is then reshaped into (BatchSize, C, Height//C\_a, Width*C\_a//C) before being incrementally scrolled along the four spatial directions. Ultimately, the output is presented as (BatchSize, Height, Width). For this study, we set C=16 and C\_a=4. The pseudo code is presented in Algorithms 1. Inserting a RollBlock between LayerNorm and FeedForward operation results in Roll-Time-mixing, as shown in Fig. 3.

\begin{table}[H]\centering
\begin{tabular}{l}
\hline
\textbf{Algorithm 1: Pytorch-like pseudo code of Roll}                                                                                                   \\ \hline
def Roll(feat, gamma=1/(1+alpha), step=ModelDepth-alpha):                                                                       \\
\qquad C, C\_a = 16, 4                                                                                                                               \\
\qquad   B, H, W = feat.shape                                                                                                                        \\
\qquad   feat = feat.reshape(B, C, H//C\_a. W//(C//C\_a))                                                                                            \\
\qquad   g = int(gamma * C)                                                                                                                          \\
\qquad   out = feat                                                                                                                                  \\
\begin{tabular}[c]{@{}l@{}}\qquad   out{[}:, 0*g:1*g, :, :{]} = torch.roll(out{[}:, 0*g:1*g, :, :{]}, \\          \qquad\qquad shifts=step, dim=3)\end{tabular}   \\
\begin{tabular}[c]{@{}l@{}}\qquad   out{[}:, 1*g:2*g, :, :{]} = torch.roll(out{[}:, 1*g:2*g, :, :{]}, \\          \qquad\qquad shifts=-step, dim=3)\end{tabular} \\
\begin{tabular}[c]{@{}l@{}}\qquad   out{[}:, 2*g:3*g, :, :{]} = torch.roll(out{[}:, 2*g:3*g, :, :{]}, \\          \qquad\qquad shifts=step, dim=2)\end{tabular}  \\
\begin{tabular}[c]{@{}l@{}}\qquad   out{[}:, 3*g:4*g, :, :{]} = torch.roll(out{[}:, 3*g:4*g, :, :{]}, \\        \qquad\qquad  shifts=-step, dim=2)\end{tabular} \\
\qquad  return out.reshape(B, H, W)                                                                                                                 \\ \hline
\end{tabular}
\end{table}

\begin{figure}[ht]
\centerline{\includegraphics{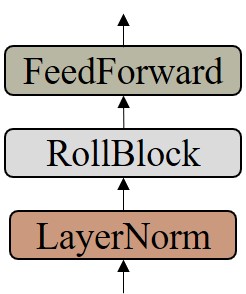}}
\caption{Roll-Time-mixing}
\label{fig3}
\end{figure}

\subsection{Hermit-Frequency-mixing}

Drawing inspiration from ActiveMLP\cite{2022ActiveMLP}, Active Token Mixer\cite{wei2023active}, and Adaptive Frequency Filters\cite{huang2023adaptive}, we introduce the concept of Hermit-Frequency-mixing in this context. The Hermit Fast Fourier Transform (FFT) is integrated before the initial data transposition, while the Inverse Real Fast Fourier Transform (IRFFT) is included after the second data transposition to establish the Hermit-Frequency-mixing, depicted in Fig. 4. The Hermit Fast Fourier Transform (HFFT) enables the model to capture frequency domain characteristics and Hermit properties of the data. The Inverse Real Fast Fourier Transform aids in restoring the data to its original domain. Notably, both transformations ensure that real inputs produce real outputs without the need for complex numbers.

\begin{figure}[ht]
\centerline{\includegraphics{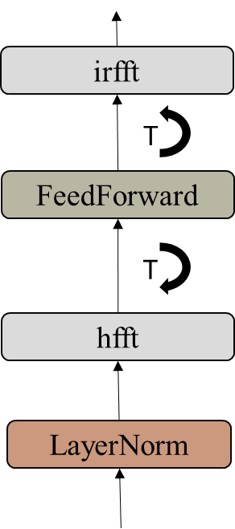}}
\caption{Hermit-Frequency-mixing}
\label{fig4}
\end{figure}

\section{Experiments}
Audio Spectrogram Mixer has proven its efficacy in audio classification tasks and comparisons with Transformers. This paper adopts ASM as a benchmark for evaluation and primarily focuses on four datasets: SpeechCommands, UrbanSound8K, CASIA Chinese Sentiment Corpus, and RAVDESS for experimentation. The first three datasets align with those used in ASM, while RAVDESS serves as the comparison target against the established state-of-the-art (SOTA) model ERANNs.All experiments were evaluated on a single NVidia RTX3080 Ti GPU.

For experimental replication purposes, here are additional details: The data shape prior to slicing was [batchsize, 600, 768] and retained this structure until reaching the MLP output layer. Specifically, 12 RH-MixerBlocks were stacked in the model architecture. Following the RH-MixerBlocks, an MLP output layer was incorporated to process the data for final predictions.

Throughout this paper, the notation "-I" signifies pre-trained models that utilize DeiT partially, where ImageNet=True, as outlined in the Audio Spectrogram Mixer paper\cite{Ji2024ASMAS}. The notation "-A" signifies pre-trained models that utilize AST partially, where AudioSet=True, as outlined in the Audio Spectrogram Transformer paper\cite{gong2021ast}.

\subsection{SpeechCommands}
Speech Commands V2 is an audio dataset with 35 classes, comprising 84,843 training samples, 9,981 validation samples, and 11,005 test samples, each lasting 1 second. Following ASM settings, the initial learning rate was set to 2.5e-4, with decay starting after the 5th epoch. Experiments were conducted for 30 epochs with ImageNet set to False.

The best model was selected based on metrics from the validation dataset, and the evaluated metrics were reported for the test set. Three different random seeds were used for the validation process of each model. The evaluation metrics encompassed accuracy (ACC, primary metric) and area under the curve (AUC, secondary metric). The mean values of the experimental results are shown in Table \ref*{table:SC}.

\begin{table}[tbp]\centering
\caption{SpeechCommand Result}
\begin{tabular}{ccccccc}
\hline
      & \textbf{ASM(\%)} & \textbf{ASM-I(\%)} & \textbf{ASM-RH(\%)} \\ \hline
v-acc & 92.92            & 94.70              & 96.62             \\
v-auc & 99.77            & 99.87              & 99.84             \\
t-acc & 91.89            & 94.01              & 96.51             \\
t-auc & 99.73            & 99.83              & 99.89             \\ \hline
\end{tabular}
\label{table:SC}
\end{table}

\subsection{UrbanSound8K}

Urbansound8K is a widely used public dataset for automatic urban environmental sound classification research. This dataset contains a total of 8732 annotated sound fragments (<=4s), including 10 categories: air conditioning sound, car honking sound, children's playing sound, dog barking, drilling sound, engine idling sound, gunfire, drill, siren sound, and street music sound. Following ASM settings, the initial learning rate was set to 2.5e-4, with decay starting after the 5th epoch. Experiments were conducted for 25 epochs with ImageNet set to False.

The best model was selected based on metrics from the validation dataset, and the evaluated metrics were reported for the test set. Three different random seeds were used for the validation process of each model. The evaluation metrics encompassed accuracy (ACC, primary metric) and area under the curve (AUC, secondary metric). The mean values of the experimental results are shown in Table \ref*{table:US8K}.

\begin{table}[ht]\centering
\caption{UrbanSound8K Result}
\begin{tabular}{ccccccc}
\hline
      & \textbf{ASM(\%)} & \textbf{ASM-I(\%)} & \textbf{ASM-RH(\%)} \\ \hline
v-acc & 88.52            & 92.30              & 96.49             \\
v-auc & 99.06            & 99.44              & 99.80             \\
t-acc & 89.51            & 91.76              & 95.80             \\
t-auc & 99.03            & 99.44              & 99.83             \\ \hline
\end{tabular}
\label{table:US8K}
\end{table}

To compare with the current state-of-the-art (SOTA) methods, we performed a 10-fold cross-validation experiment on the UrbanSound 8K dataset. The average and best results are presented in Table \ref*{table:US8K 10-fold} for reference.

\begin{table}[ht]\centering
	\caption{UrbanSound8K 10-fold Result}
	\begin{tabular}{cccc}
	\hline
				 & \textbf{Avg acc(\%)} & \textbf{Best acc(\%)} \\ \hline
		EAT-M    & 90                   & -                     \\
		ASM-RH-I & 97.96                & 98.63                 \\ \hline
	\end{tabular}
	\label{table:US8K 10-fold}
	\end{table}

\subsection{CASIA Chinese Sentiment Corpus}

CASIA Chinese Sentiment Corpus covers speech material recorded by four professional pronouncers, including six emotions: angry, happy, fear, sad, surprised and neutral. A total of 9600 speech samples with different pronunciations are included. Following ASM settings, the initial learning rate was set to 2.5e-4, with decay starting after the 5th epoch. Experiments were conducted for 25 epochs with ImageNet set to False.

The best model was selected based on metrics from the validation dataset, and the evaluated metrics were reported for the test set. Three different random seeds were used for the validation process of each model. The evaluation metrics encompassed accuracy (ACC, primary metric) and area under the curve (AUC, secondary metric). The mean values of the experimental results are shown in Table \ref*{table:CASIA}.

\begin{table}[H]\centering
\caption{CASIA Result}
\begin{tabular}{ccccccc}
\hline
      & \textbf{ASM(\%)} & \textbf{ASM-I(\%)} & \textbf{ASM-RH(\%)} \\ \hline
v-acc & 90.25            & 92.47              & 93.15             \\
v-auc & 99.48            & 99.62              & 99.60             \\
t-acc & 89.97            & 91.76              & 92.19             \\
t-auc & 99.43            & 99.45              & 99.45             \\ \hline
\end{tabular}
\label{table:CASIA}
\end{table}

\subsection{RAVDESS}

RAVDESS (Ryerson Audio-Visual Database of Emotional Speech and Song) is an emotional speech and song database contributed by participants from Ryerson University, Canada. This dataset includes recordings of audio clips featuring various emotional states. It comprises 24 performers, evenly split between males and females, who deliver speech and songs depicting emotions such as happiness, neutrality, sadness, anger, surprise, fear, disgust, and calmness. In this section, we employ a 10-fold cross-validation method and consider the best performance achieved across the folds as the final result. The result is in Table \ref*{table:RAVDESS}.

\begin{table}[ht]\centering
\caption{RAVDESS Result}
\begin{tabular}{ccccccc}
\hline
                   & \textbf{acc(\%)} & \textbf{auc(\%)} \\ \hline
\textbf{ERANN-1-3} & 73.1             & \textbf{-}       \\
\textbf{ERANN-0-4} & 74.8             & \textbf{-}       \\
\textbf{ERANN-1-4} & 74.1             & \textbf{-}       \\ \hline
\textbf{ASM-RH-A}      & 75.4            & 94.98            \\ \hline
\end{tabular}
\label{table:RAVDESS}
\end{table}

\subsection{Ablation Study}
To assess the effectiveness of Roll-Time-mixing and Hermit-Frequency-mixing, ablation experiments were conducted on the UrbanSound8K dataset. Channel-mixing and Tokens-mixing in the MLP-Mixer were implemented as alternatives to Roll-Time-mixing and Hermit-Frequency-mixing, respectively. The experiment maintained consistency by utilizing the same three random seeds, ensuring reproducibility and reliability. The results were averaged and are presented in Table \ref*{table:Ablation of US8K} for a comprehensive comparison and analysis. In the table, "-H" represents ASM-RH with Hermit-Frequency-mixing only, while "-R" signifies ASM-RH with Roll-Time-mixing only.

The best model was chosen based on the metrics obtained from the validation dataset, and the performance metrics were then reported for the test set. The evaluation metrics encompassed accuracy (ACC, primary metric) and area under the curve (AUC, secondary metric).

\begin{table}[htbp]\centering
\caption{UrbanSound8K Ablation Experiment Result}
\begin{tabular}{ccccccc}
\hline
      & \textbf{ASM-RH(\%)} & \textbf{ASM-H(\%)} & \textbf{ASM-R(\%)} \\ \hline
v-acc & 96.49             & 95.61               & 96.19               \\
v-auc & 99.80             & 99.76               & 99.77               \\
t-acc & 95.80             & 95.19               & 95.04               \\
t-auc & 99.83             & 99.75               & 99.79               \\ \hline
\end{tabular}
\label{table:Ablation of US8K}
\end{table}

\subsection{Discussion}

In this paper, we introduce the Audio Spectrogram Mixer with Roll-Time and Hermit FFT model, which mix time-domain and frequency-domain information. We compare this model with the Audio Spectrogram Mixer\cite{Ji2024ASMAS}, which integrates Mixer into audio classification tasks, across three datasets: SpeechCommands, UrbanSound8K, and the CASIA Chinese sentiment corpus. The experimental findings reveal that, even without pre-training or additional information, the Audio Spectrogram Mixer with Roll-Time and Hermit FFT model outperforms the Audio Spectrogram Mixer significantly. Indeed, the results demonstrate that the performance of the Audio Spectrogram Mixer with Roll-Time and Hermit FFT model surpasses that of the Audio Spectrogram Mixer, even when the latter is enhanced with visual information. This underscores the effectiveness and promising outcomes of the proposed ASM-RH model.

Furthermore, our evaluation on the RAVDESS dataset illustrated that the Audio Spectrogram Mixer with Roll-Time and Hermit FFT model outperforms the state-of-the-art (SOTA) ERANN models, showcasing its superior performance in audio classification tasks.

Lastly, ablation experiments conducted on the UrbanSound 8K dataset validated that Roll-Time-mixing and Hermit-Frequency-mixing effectively capture both time-domain and frequency-domain information, further reinforcing the efficacy of Mixer in audio data processing and classification tasks.

\section{Mixer is more than just a model}

In this paper, we introduce two novel structures, Roll-Time-mixing and Hermit-Frequency-mixing, designed to capture time-domain and frequency-domain information, respectively. These structures are integrated into Mixer to create the ASM-RH model for audio classification tasks. Experimental results validate the effectiveness of these structures and the capability of ASM-RH.

The audio categorization task serves as a demonstration of Mixer's potential, but the essence of this paper is to urge researchers to delve deeper into the study of MLP and Mixer. As previously mentioned, Mixer transcends being merely a model structure; it represents a mind and approach to handling and interpreting data. Just as CNNs blend information from local and global viewpoints, and spatio-temporal models\cite{Liu2024HowCL} merge temporal and spatial information, they can be viewed as a form of Mixer. Just as MLP-Mixer\cite{2021MLP} looks at visual data from Channel and Tokens perspectives, and ASM-RH looks at spectrogram data from time and frequency domain perspectives, we hope that more researchers will develop high-quality models that capture and MIX information from more perspectives. All are Mixers.

\bibliographystyle{IEEEbib}
\bibliography{refs}
\vspace{12pt}
\end{document}